\documentclass[pdflatex,sn-mathphys-num]{sn-jnl}%

\usepackage{graphicx}
\usepackage{multirow}
\usepackage{amsmath,amssymb,amsfonts}
\usepackage{amsthm}
\usepackage{mathrsfs}
\usepackage[title]{appendix}
\usepackage{xcolor}
\usepackage{textcomp}
\usepackage{manyfoot}
\usepackage{booktabs}
\usepackage{algorithm}
\usepackage{algorithmicx}
\usepackage{algpseudocode}
\usepackage{listings}
\usepackage{float}
\usepackage{booktabs}
\usepackage{multirow}
\usepackage{lineno}

\theoremstyle{thmstyleone}

\theoremstyle{thmstyletwo}

\theoremstyle{thmstylethree}

\begin{document}

\title[Article Title]{
Agentic evolution of physically constrained foundation models
}

\author*[1]{\fnm{Jiangwei} \sur{Zhang}}\email{zhangjiangwei@iie.ac.cn}
\equalcont{These authors contributed equally to this work.}

\author[1,2]{\fnm{Wen} \sur{Sun}}
\equalcont{These authors contributed equally to this work.}

\author[3]{\fnm{Chong} \sur{Wang}}
\equalcont{These authors contributed equally to this work.}

\author[3]{\fnm{Shiyao} \sur{Li}}
\equalcont{These authors contributed equally to this work.}

\author[1,2]{\fnm{Cheng} \sur{Che}}

\author[1,2]{\fnm{Chunjing} \sur{Han}}

\author[1,2]{\fnm{Dan} \sur{Meng}}

\author[3]{\fnm{Jian} \sur{Yang}}

\author*[3]{\fnm{Yu} \sur{Wang}}\email{yu-wang@mail.tsinghua.edu.cn}

\author*[1,2]{\fnm{Rui} \sur{Hou}}\email{hourui@iie.ac.cn}

\affil[1]{\orgdiv{State Key Laboratory of Cyberspace Security Defense}, \orgname{Institute of Information Engineering, Chinese Academy of Sciences}, \orgaddress{\city{Beijing}, \postcode{100084}, \country{China}}}

\affil[2]{\orgname{University of Chinese Academy of Sciences}, \orgaddress{\city{Beijing}, \postcode{101408}, \country{China}}}

\affil[3]{\orgdiv{Department of Electronic Engineering}, \orgname{Tsinghua University}, \orgaddress{\city{Beijing}, \postcode{100084}, \country{China}}}

\abstract{

Artificial intelligence increasingly drives automated scientific discovery, yet contemporary generalist agents lack physical grounding, frequently hallucinating hardware-incompatible designs. Here, we present a physically grounded, multi-agent discovery engine that autonomously architects hardware-compliant computing systems. Anchored by an Evolutionary Knowledge Graph structuring past scientific innovations, the framework extracts an ``algorithmic Chain-of-Thought'' to transform blind stochastic search into directed structural evolution. Applied to the extreme testbed of foundation model deployment, the engine evolved two hardware-aware compression methodologies surpassing human-engineered heuristics: Q-Enhance mitigates long-context accuracy loss in dense models, and MoE-Salient-AQ outperforms state-of-the-art manual sparse Mixture-of-Experts designs by 3.7\% at sub-3-bit regimes. Utilizing a bandwidth-efficient Sensitivity Profile, we successfully deployed a massive 235-billion-parameter model onto a constrained dual-A100 server, reducing memory requirements by 75\% with a marginal 0.64\% accuracy degradation. By transforming unconstrained combinatorial search into knowledge-driven autonomy, this establishes a scalable hardware-software co-design paradigm for machine-driven discovery within strict physical boundaries.

}


\maketitle

\section*{Main Text}\label{sec1}

Artificial intelligence (AI) has emerged as a transformative force across scientific disciplines, driving discoveries from protein folding to materials design \cite{r1, r2, r6, AlphaFold, m12, Degrave2022Magnetic, gao2026autonomous, merchant2023scaling}. However, while the machine learning community is experiencing a paradigm shift toward AI-driven autonomous research \cite{lu2026aiscientist, r31, karpathy2026autoresearch, eureka, bu2026empowering}, translating this autonomy into physical, hardware-aware system design remains a formidable frontier \cite{r3, r9, m10, m11}. Current automated methodologies generally bifurcate into two inadequate paradigms. On one hand, contemporary autonomous agents predominantly operate within high-level software environments. Lacking true physical grounding, they rely on fragmented semantic retrieval and implicit parametric knowledge, making them highly prone to ``physical hallucinations''—frequently proposing abstract algorithmic structures that violate immutable physical constraints \cite{r35, r36, r37, kalai2026evaluating}. On the other hand, traditional automation tools avoid these hallucinations but remain rigidly restricted to combinatorial search within predefined, narrow design spaces \cite{m5, m6, PostNAS}. Both paradigms share a fundamental flaw: they fail to explicitly model the structural evolution of prior designs. Without a structured historical repository mapping how past optimizations successfully navigated strict hardware limits, these systems remain trapped in redundant trial-and-error, unable to autonomously architect physically compliant computing systems.

Here we present an agent-driven discovery engine that bridges this gap by architecting its own hardware-aware algorithmic evolution. Driven by a heterogeneous multi-agent framework, the system is anchored by an Evolutionary Knowledge Graph (EKG). Constructed from 164 mainstream large language model (LLM) compression methods, this EKG serves as a dynamically updatable, machine-readable repository of scientific innovations \cite{r13, r15, r40, r39}. By formulating an ``algorithmic Chain-of-Thought'' (CoT), the framework retraces and mutates validated reasoning paths, seamlessly transforming abstract ideas into executable algorithmic blueprints without falling back on blind stochastic search. To navigate the high-dimensional and non-differentiable open-ended design space, rigorous AI-driven peer review protocols independently filter hundreds of generated hypotheses. Empirical evaluations confirm that this multi-agent synergy effectively mitigates the hardware-incompatible failures inherent to unconstrained generalist agents, yielding topological blueprints that objectively match the logical rigor of expert-level peer-reviewed literature.

To rigorously validate this framework's capacity to navigate intractable combinatorial explosions governed by strict physical boundaries, we applied it to one of the most critical bottlenecks in contemporary computing: the hardware-aware compression of foundation models. The exponential scaling of foundation model parameters now severely outpaces advancements in memory capacity and bandwidth \cite{r16}. Bridging this hardware--software gap requires highly efficient compression strategies. However, the rapid evolution toward complex computational paradigms (e.g., long-context processing and sparse routing) has fundamentally altered the nature of this optimization. Dynamically allocating computational precision and memory resources across these vast architectures creates an intractable physical search space. Because this complexity far surpasses the limits of human intuition, traditional heuristic-driven compression strategies, which often consume months of manual trial-and-error, have become severely inadequate~\cite{kim2025efficient}. Consequently, relying on human experts to manually redesign optimization algorithms is an inherently unscalable paradigm.

Operating within this extreme physical testbed, the agentic engine evolved two robust compression methods that redefine the accuracy--efficiency Pareto frontier by directly resolving two of the field's most critical bottlenecks: the memory explosion of long-sequence processing and the quantization instability inherent to Mixture-of-Experts (MoE) models. For dense foundation models, it engineered Q-Enhance, an outlier-aware mechanism that dynamically reallocates precision to prevent performance collapse at sequence lengths up to 128k tokens \cite{ATOM, KVQuant}. For sparse architectures, it evolved MoE-Salient-AQ, which establishes an expert-granularity quantization topology. By coupling activation-driven precision allocation with adaptive low-rank error compensation, it maintains reasoning capability under extreme compression, outperforming human-designed state-of-the-art baselines by 3.7\% at sub-3-bit regimes~\cite{SliM-LLM, DWR}.

Bridging theoretical discovery with physical deployment requires closing the loop between algorithmic blueprints and strict hardware limits. By leveraging a bandwidth-efficient Sensitivity Profile to act as a direct hardware feedback mechanism, the system dynamically tailors its evolutionary strategies to specific physical constraints. This enables the direct deployment of 8B--parameter models on a single consumer-grade GPU for real-time edge applications. Furthermore, it successfully compresses massive 235B-parameter foundation models onto dual-A100 servers for memory-bound cloud inference, reducing hardware requirements by 75\% (from 438 GB to 108 GB) with a marginal 0.64\% accuracy degradation. By transforming hardware-aware design from unconstrained search into knowledge-driven autonomy, this framework establishes a scalable paradigm for the automated hardware--software co-design of computing systems within strict physical boundaries.

\section*{Framework and Algorithmic Memory}

\begin{figure}[htb]
\centering
\includegraphics[width=\textwidth]{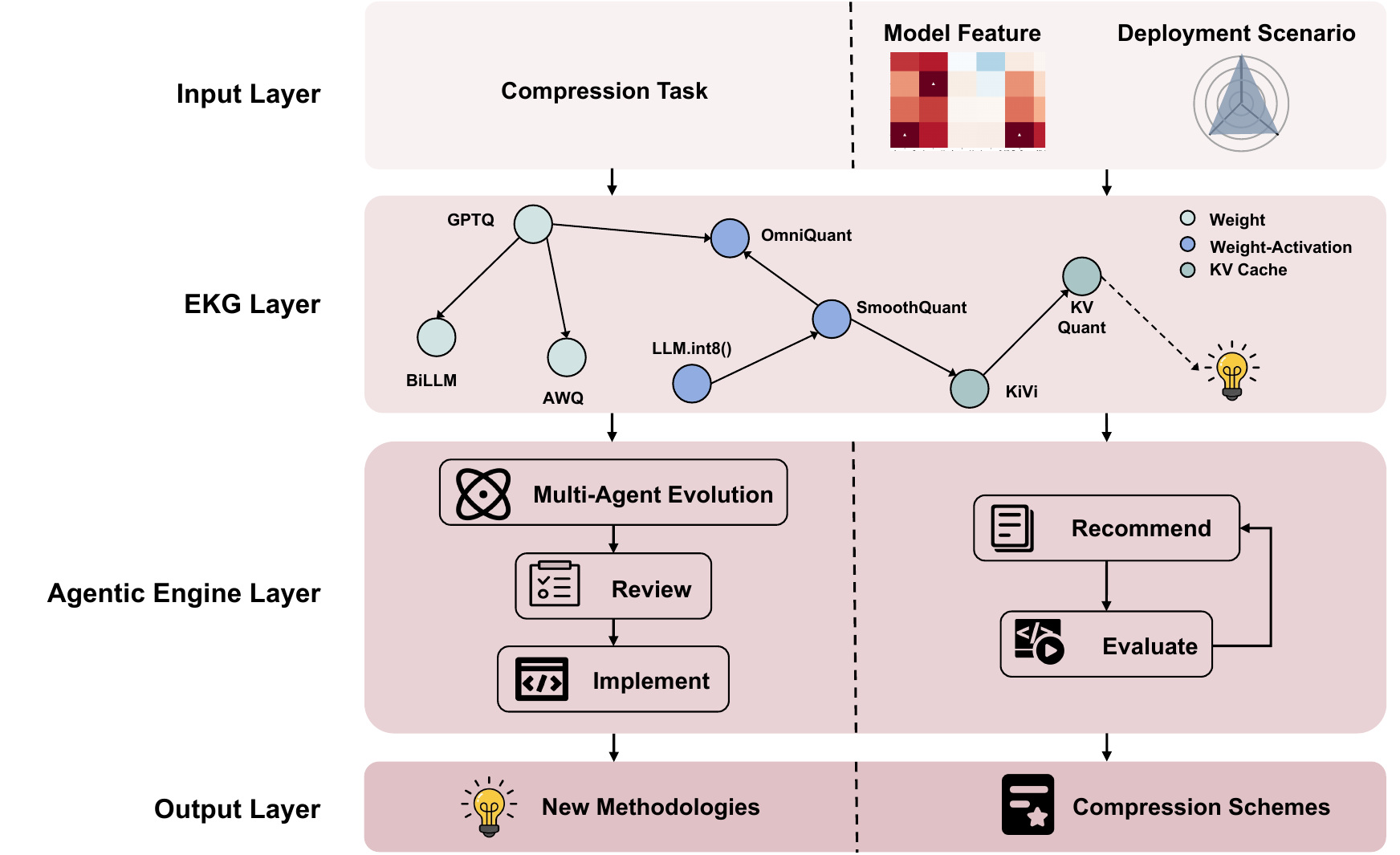}
\caption{\textbf{Architecture of the autonomous closed-loop discovery engine.} The framework is structured across four functional layers, integrating an autonomous algorithmic evolution cycle on the left with a constraint-aware hardware instantiation flow on the right. \textbf{Input Layer:} Processes abstract compression objectives for global architectural evolution, alongside hardware deployment constraints and lightweight model profiling metrics for local hardware mapping. \textbf{EKG Layer:} Serving as the system's foundational memory, it organizes historical optimization trajectories, including weight, activation, and KV cache quantization paradigms. It provides structured evolutionary priors to guide agentic reasoning. \textbf{Agentic Engine Layer:} A heterogeneous ensemble of specialized agents operates directly atop the EKG. The left pathway autonomously hypothesizes novel methodologies, conducts rigorous AI peer-reviews, and executes physical implementations. Concurrently, the right pathway features a hardware-aware calibration loop that evaluates candidate configurations to determine the optimal deployment strategy. \textbf{Output Layer:} Yields validated algorithmic methodologies and hardware-tailored compression schemes. Successful derivations establish a deterministic feedback loop, indicated by the dashed integration path in the EKG, dynamically updating the graph to facilitate continuous, self-improving scientific discovery.
}\label{fig1}
\end{figure}

As illustrated in Figure~\ref{fig1}, the agent-driven discovery engine is conceptualized as a unified four-layer architecture that integrates a generative algorithmic evolution flow and a constraint-aware hardware instantiation loop. The pipeline initiates at the Input Layer, which processes either high-level compression tasks to drive algorithmic discovery, or bandwidth-efficient Sensitivity Profiles tailored for strict deployment scenarios. These inputs query the EKG Layer, a shared, dynamic memory structure that serves as the system's foundational knowledge base (Extended Data Fig.~\ref{fig:ED1}). This graph structures the historical trajectories of optimization into two distinct representational layers (see Methods for construction details). Operating atop this memory is the Agentic Engine Layer, where a collaborative ensemble of specialized agents autonomously hypothesizes, rigorously reviews, and computationally implements novel methodologies, while a parallel iterative loop recommends and evaluates optimal, hardware-aware deployment strategies. Finally, the Output Layer yields either validated methodologies or locally tailored physical mappings.

At the Macro-Topology level of the EKG Layer, the network organizes algorithms into a branching relational structure, mapping the logical dependencies between isolated methodologies. It visualizes the divergence from a root ancestor into specialized evolutionary branches, tracking the progression within weight-only optimizations such as the evolution from GPTQ~\cite{gptq} to AWQ~\cite{r13}, alongside parallel paths for joint weight-activation methods like SmoothQuant~\cite{r15} and mechanisms for key-value (KV) cache management. This topological separation allows the discovery engine to navigate algorithmic divergence, pinpointing complementary mechanisms across distinct methodological families that are prime for integration.

\begin{figure}[H]
\centering
\includegraphics[width=\textwidth]{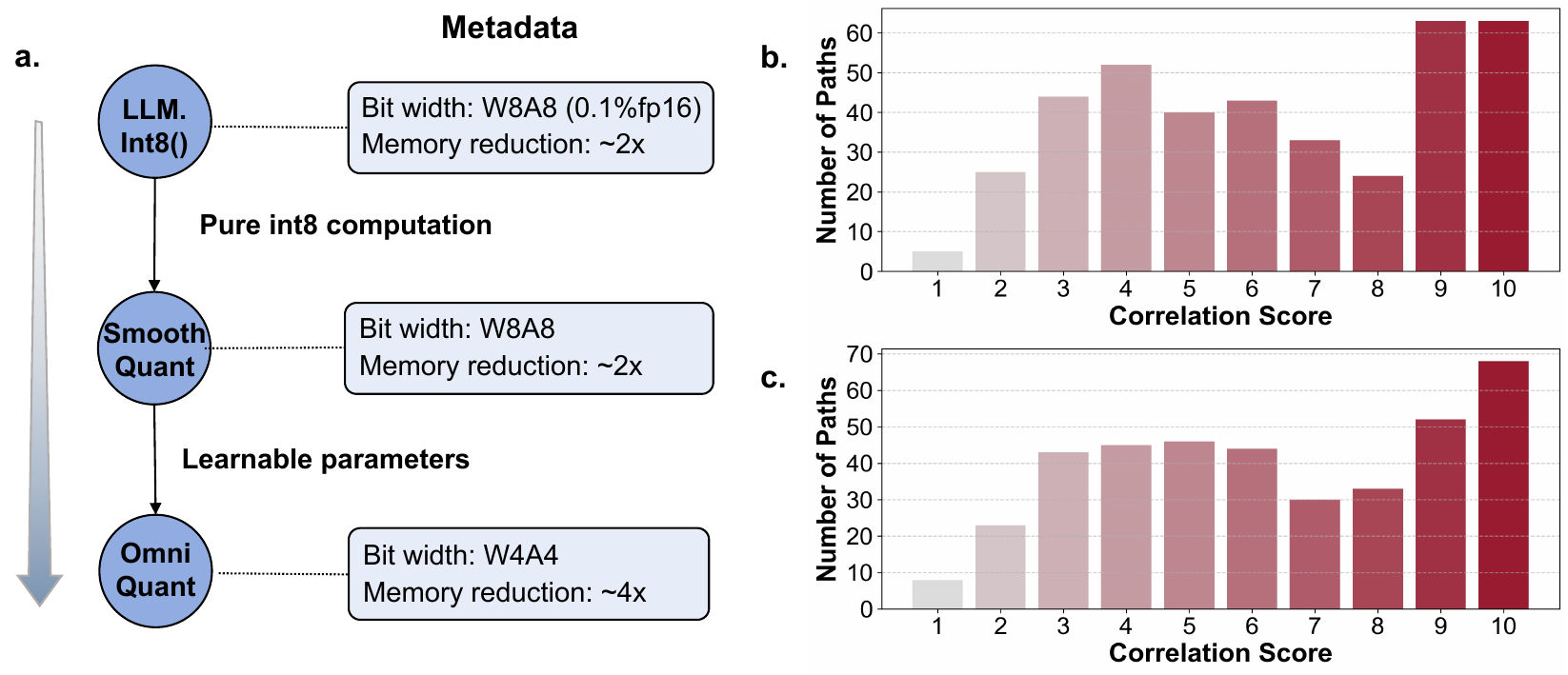}
\caption{\textbf{Granular trajectory tracking and algorithmic evaluation within the EKG.} \textbf{a,} Path tracking and metadata encapsulation. The EKG explicitly tracks algorithmic transitions along specific optimization paths (e.g., the progression from LLM.int8() \cite{dettmers2022gpt3} to SmoothQuant \cite{r15} to OmniQuant \cite{omniquant}). Each node maintains a metadata record of key system attributes, such as numerical precision (bit-width) and memory reduction ratios. Directed edges annotate the specific structural transition mechanisms (e.g., the introduction of learnable parameters), enabling precise quantitative tracking of methodological advancements. \textbf{b, c,} Systematic trajectory traversal and evaluation. Histograms illustrate the autonomous engine's comprehensive traversal and correlation scoring (1--10 scale) of historical paths, tailored for dense (\textbf{b}) and sparse (\textbf{c}) compression tasks. The scoring distributions cluster strongly toward high-relevance trajectories, demonstrating the system's robust discriminative capability. The engine effectively filters out architecturally incompatible methodologies (low scores) while isolating highly complementary trajectories (high scores) for subsequent directed evolution.}
\label{fig2}
\end{figure}

At the Micro-Dynamics level (Figure~\ref{fig2}a), the EKG functions as a sequence of evolutionary checkpoints. Each node maintains a structured metadata record that explicitly records system attributes, including numerical precision formats like W8A8 (8-bit weight, 8-bit activation), memory hardware compatibility, and the specific transitional mechanisms, such as the introduction of learnable parameters, that facilitate the algorithmic progression~\cite{dettmers2022gpt3, r15, omniquant}.

By explicitly tracking these methodological advancements, the autonomous engine conducts a systematic traversal and scoring of the evolutionary hypothesis space (Figures~\ref{fig2}b and \ref{fig2}c). This task-specific evaluation yields a highly polarized scoring distribution clustered strongly toward high-relevance trajectories. Such polarization demonstrates the system's robust discriminative capability: it effectively filters out incompatible methods and isolates complementary paths, focusing computational resources exclusively on viable blueprints. Furthermore, negative control evaluations on conceptually orthogonal domains yielded uniformly low correlation scores, confirming that this matching is strictly task-driven and highly resistant to retrieval hallucinations (Extended Data Fig.~\ref{fig:ED5}).

\section*{Agent-Driven Algorithmic Evolution}

\begin{figure}[H]
\centering
\includegraphics[width=\textwidth]{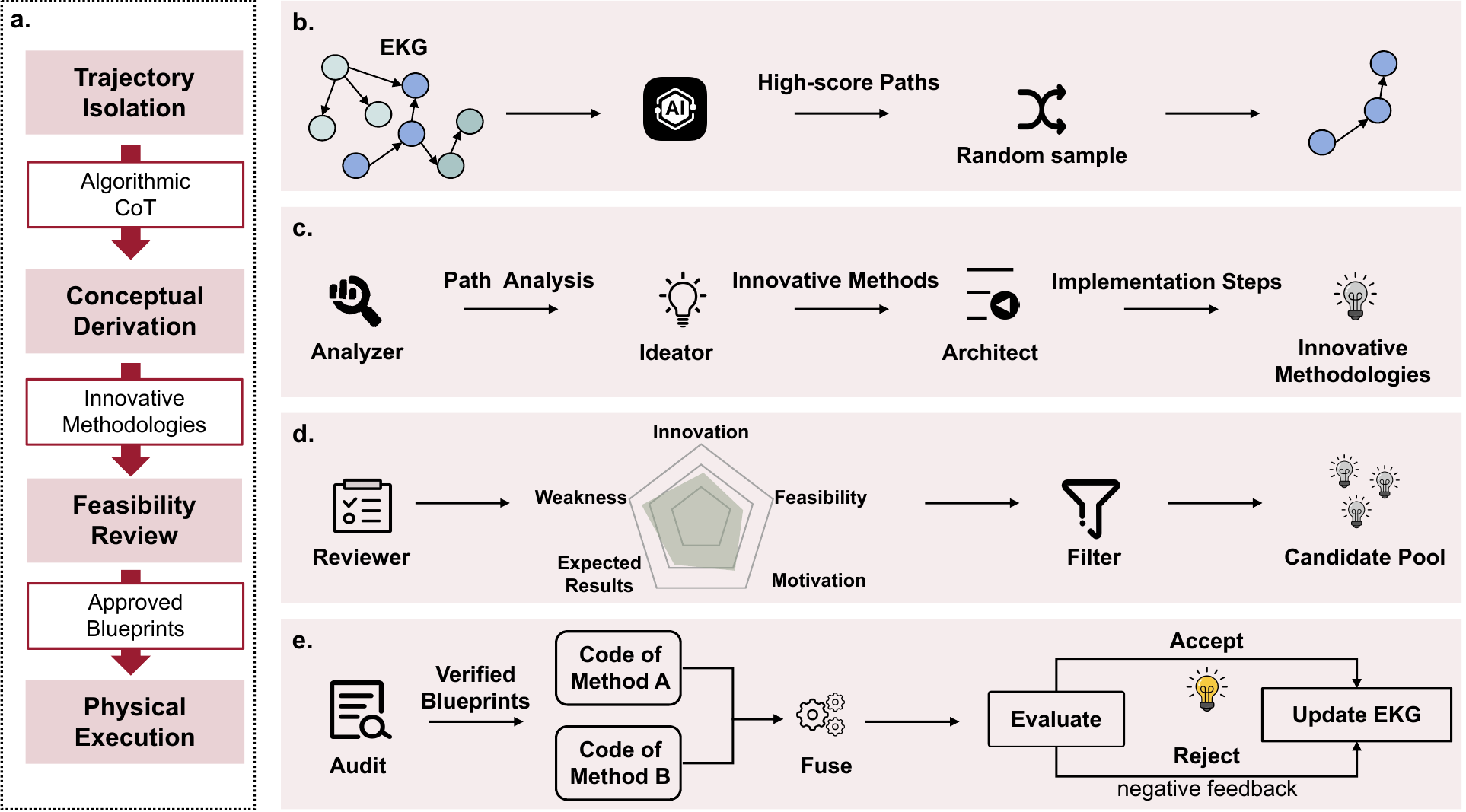}
\caption{\textbf{Autonomous algorithmic evolution via multi-agent collaboration.} \textbf{a,} Overarching evolutionary pipeline. The framework transforms high-level compression tasks into executable code through four sequential stages. \textbf{b,} Trajectory isolation. The system traverses the EKG to isolate high-potential optimization paths, outputting a collection of high-score trajectories to serve as an algorithmic CoT. \textbf{c,} Conceptual derivation. A triad of foundation models (Analyzer, Ideator, and Architect) performs rigorous path analysis, proposes innovative methods, and formalizes detailed implementation steps. \textbf{d,} Feasibility review. An autonomous AI peer-review protocol evaluates candidates against multidimensional criteria (visualized via a radar chart scoring weakness, expected results, motivation, feasibility, and innovation), indexing only passing designs into a candidate pool. \textbf{e,} Physical execution. Verified blueprints, which have undergone stringent pre-implementation logic auditing, are translated into reality by fusing reference code snippets (e.g., Code of Method A and B). Candidates are evaluated empirically; successful deployments that redefine the Pareto frontier are integrated to continuously enrich the EKG, while empirical failures provide negative feedback to refine future navigational strategies.}
\label{fig3}
\end{figure}

As delineated by the overarching evolutionary pipeline (Figure~\ref{fig3}a), the generation of novel methodologies begins by isolating the highest-potential historical trajectories (Figure~\ref{fig3}b). To balance exploration with exploitation within this filtered candidate pool, the system stochastically samples a high-scoring optimization path. This selected path serves as an algorithmic CoT context, guiding a heterogeneous ensemble of foundation models tailored for distinct analytical tasks.

This multi-agent collaboration protocol (Figure~\ref{fig3}c) transforms the high-level trajectory context into meticulously vetted algorithmic blueprints. First, the Analyzer agent dissects the chosen source trajectory, producing a structured diagnostic of the internal mechanisms underlying previous advancements. This informs the Ideator agent, which employs divergent reasoning to propose conceptual hypotheses optimized for the new compression objective. These proposals are then translated by the Architect agent into detailed algorithmic blueprints, specifying the mathematical formulations and execution logic (see Methods: Heterogeneous Multi-Agent Evolution).

Before physical implementation, these theoretical blueprints undergo an autonomous AI peer review (Figure~\ref{fig3}d). Applying evaluation criteria analogous to standard academic review, the Reviewer agent objectively scores each candidate's feasibility and innovation against strict logical and hardware bounds (Extended Data Fig.~\ref{fig:ED2}a). Only blueprints that exceed the acceptance threshold are deposited into a viable candidate pool (see Methods: Automated AI Peer Review Protocol). 

The final phase translates these theoretical strategies into executable pipelines (Figure~\ref{fig3}e). Prior to explicit code synthesis, selected designs from the candidate pool undergo a stringent logic auditing phase to preemptively resolve latent theoretical and physical flaws. The refined candidates are then instantiated via autonomous algorithmic integration, incorporating open-source operators from the most relevant reference implementations retrieved from the EKG. The resulting executable candidates are empirically benchmarked against established human-designed baselines. This iterative process establishes a bidirectional scientific feedback loop (Extended Data Fig.~\ref{fig:ED2}b): successful innovations that advance the Pareto frontier are integrated into the EKG as new positive nodes. Conversely, empirical failures are logged as trajectory-level negative feedback to dynamically refine the system's future navigational strategies.

\section*{Hardware-Aware Deployment Strategy}

To bridge global algorithmic discovery with localized physical execution, the system employs a bandwidth-efficient instantiation workflow. Central to this process is the local generation of a Sensitivity Profile (Extended Data Fig.~\ref{fig:ED3}). By condensing the foundation model's Architecture Block and outlier-driven Numerics Block into a structured JSON payload, this protocol drastically reduces the transmission overhead from terabytes of raw weights to a sub-megabyte payload of statistical abstractions. 

By projecting this profile against specific hardware constraints, the engine queries the EKG to identify the optimal path. The system subsequently refines deployment parameters through an automated calibration process. This ensures robust inference deployment across diverse consumer-grade and cloud environments (see Methods: Constraint-Aware Hardware Instantiation).

\section*{Results}

By navigating the EKG under strict performance constraints, the engine autonomously discovered two novel algorithms: Q-Enhance for dense LLMs and MoE-Salient-AQ for sparse architectures. Here, we first evaluate these novel methods. We then benchmark their logical and structural quality against both human-authored baselines and contemporary generalist agents. Finally, we demonstrate the engine's hardware-aware deployment capabilities under diverse VRAM constraints, followed by rigorous ablation studies isolating the core mechanisms driving this autonomous capability (see Methods: Experimental Setup and Evaluation Protocol). 

\begin{figure}[htb]
\centering
\includegraphics[width=\textwidth]{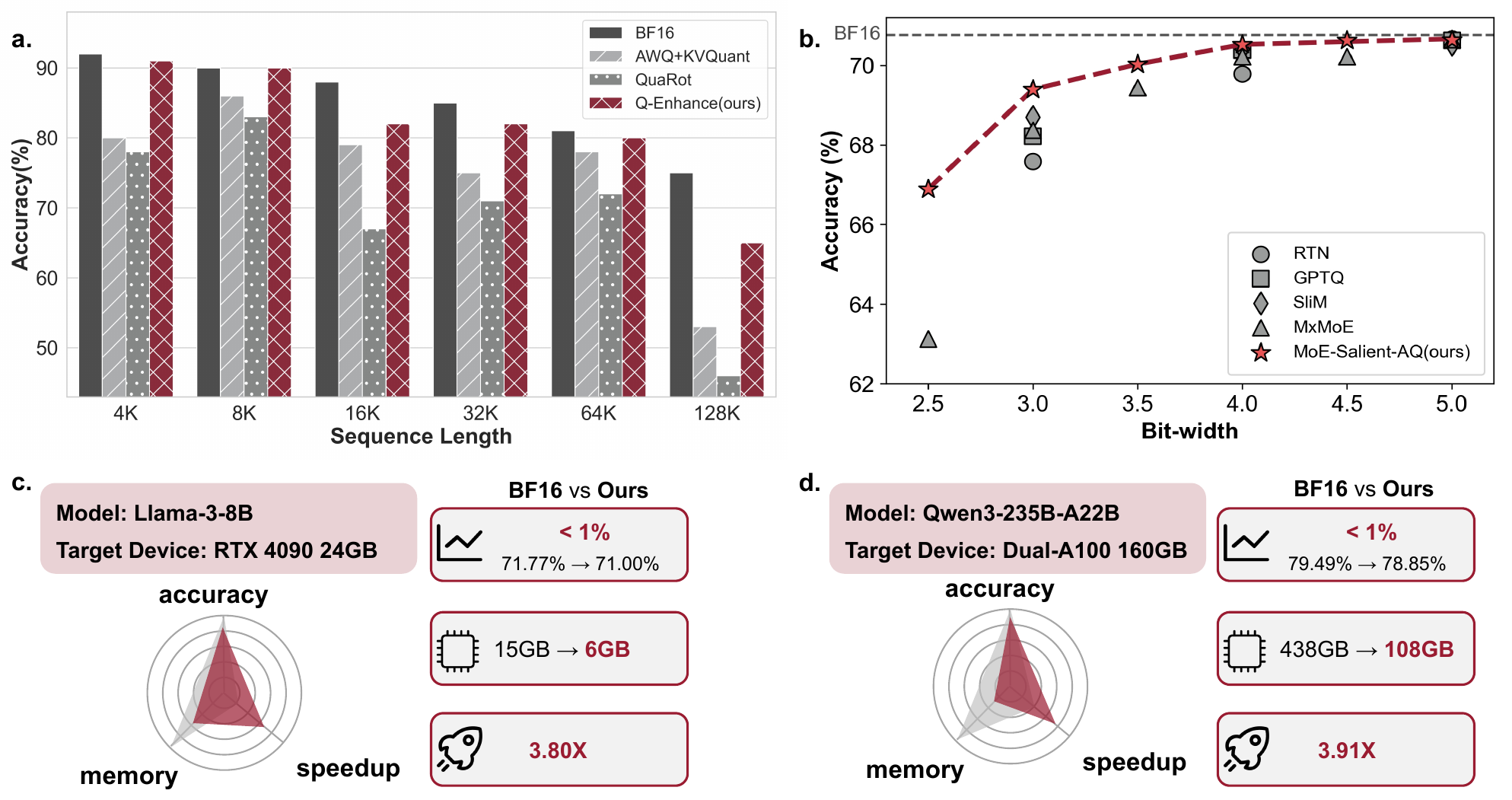}
\caption{\textbf{Algorithmic performance frontiers and hardware-aware physical instantiation of optimal compression strategies.} \textbf{a,} Long-context inference stability for dense models. The autonomously derived Q-Enhance framework effectively mitigates performance collapse at extreme context limits (up to 128k) compared to conventional baselines. By dynamically reallocating the error budget between weights and KV caches, it maintains robust reasoning capabilities. \textbf{b,} Accuracy–efficiency Pareto frontier for sparse architectures. The MoE-Salient-AQ algorithm consistently outperforms state-of-the-art baselines, a trend particularly pronounced in extreme sub-3-bit compression regimes, validating the discovered expert-level adaptive compensation. \textbf{c, d,} Scenario-driven hardware instantiation via radar charts comparing agent-recommended configurations against generic baselines. Crucially, the autonomously discovered strategies strictly bound the accuracy degradation to less than 1\% across both deployment scales. \textbf{c,} Latency-critical consumer scenario using Llama-3-8B on an RTX 4090 (24 GB), where the system achieves a substantial reduction in latency while compressing the physical memory footprint from 15 GB to 6 GB. \textbf{d,} Memory-bound cloud inference for a 235B-parameter sparse model on a dual-A100 (160 GB) server. Notably, the engine compresses the massive 438 GB theoretical workload down to 108 GB, enabling physical deployment with a marginal 0.64\% accuracy degradation.}\label{fig4}
\end{figure}

To overcome the long-context memory bottleneck in dense models, the engine evolved Q-Enhance, a joint quantization framework that dynamically reallocates bit-widths between model weights and KV caches. Unlike conventional isolated compression heuristics, this context-aware calibration adheres to a strict 4-bit equivalent memory budget while adaptively scaling to extended sequence lengths. Empirically, Q-Enhance effectively mitigates the catastrophic accuracy degradation typical of current state-of-the-art baselines (e.g., QuaRot~\cite{quarot}) under memory-constrained long-context regimes. Evaluated on Llama-3.1-8B-Instruct (Figure~\ref{fig4}a), it substantially stabilizes the performance trajectory across context windows up to 128k tokens. This exceptional long-context robustness is further corroborated on the Qwen3 family (Extended Data Table~\ref{tab:qwen3_qenhance}), confirming that the framework preserves foundational reasoning capabilities across diverse dense architectures and consistently outperforms static baselines under equivalent memory constraints.

For sparse architectures, the engine autonomously derived MoE-Salient-AQ, a post-training framework coupling expert-granularity bit-allocation with an adaptive error recovery mechanism. Unlike human-engineered state-of-the-art methods (e.g., MxMoE~\cite{mxmoe}) that rely on increasingly complex, fine-grained schemes, this framework demonstrates that precisely calibrated expert-level allocation effectively prevents representational collapse. Empirically, MoE-Salient-AQ substantially surpasses existing baselines across the accuracy--efficiency Pareto frontier. Specifically, at the extreme 2.5-bit regime, it outperforms the leading manual baseline by 3.7\% on Deepseek-V2-Lite (Figure~\ref{fig4}b) and delivers a consistent 3.3\% improvement (63.88\% vs. 60.53\%) on Qwen1.5-MoE-A2.7B (Extended Data Table~\ref{tab:moe_compression}). These results confirm the system's ability to preserve robust reasoning capabilities at extreme compression ratios where conventional manual heuristics falter.

To further evaluate the broader utility of our generated architectures across diverse provenances before physical implementation, we conducted an exploratory LLM-simulated test (Extended Data Fig.~\ref{fig:ED4}). For this evaluation, the ranking spectrum is uniformly partitioned into four equi-proportional tiers (Tier 1 to Tier 4), where Tier 1 represents the highest caliber of innovation. Our designs were anonymously ranked against state-of-the-art autonomous agents (e.g., The AI Scientist~\cite{lu2026aiscientist}), arXiv preprints, and human-authored publications from standard and top-tier venues. The results reveal a striking distribution: our system entirely bypassed the lower-quality tiers (Tiers 3 and 4) that dominate both The AI Scientist and standard venues, while avoiding the high variance of preprints. With 100\% of our blueprints placing in the upper two tiers and 40\% reaching Tier 1, our output distribution closely aligns with the rigorous standards of top-tier human expert publications.

Translating these candidate methodologies into physical reality, the deployment workflow leverages the model sensitivity profile. Governed by a strict $1\%$ accuracy degradation tolerance evaluated on task-specific benchmarks (see Methods: Experimental Setup and Evaluation Protocol), the system dynamically tailors its quantization strategies across diverse hardware environments. 

Latency-Critical Consumer Environments (e.g., real-time personal assistants, Figure~\ref{fig4}c): Deployed on a single NVIDIA RTX 4090 (24 GB VRAM) for Llama-3-8B, the system analyzed the profile and converged on an aggressive 4-bit weight-only compression scheme. This autonomously calibrated strategy delivered a substantial $3.80\times$ theoretical speedup in token generation speed. Simultaneously, it reduced the physical VRAM footprint from 15 GB to 6 GB while incurring a marginal 0.77\% accuracy degradation.

Memory-Bound Cloud Inference (e.g., large-scale foundation models, Figure~\ref{fig4}d): Faced with the 438\,GB memory requirement of the uncompressed BF16 baseline, the engine dynamically shifted its optimization objective to extreme footprint reduction---compressing the intensive Qwen3-235B-A22B model to just 108\,GB for deployment on a constrained dual-A100 server. Guided by the sensitivity profile, it achieved this by prescribing a rigorous 3.5-bit weight-only schema, which dynamically assigned mixed INT3 and INT4 precision according to expert activation salience. Despite a $4\times$ reduction in hardware footprint, the system preserved robust reasoning capabilities with a mere 0.64\% accuracy drop (79.49\% to 78.85\%), safely within the 1\% deployment tolerance.

\begin{figure}[htb]
\centering
\includegraphics[width=0.9\textwidth]{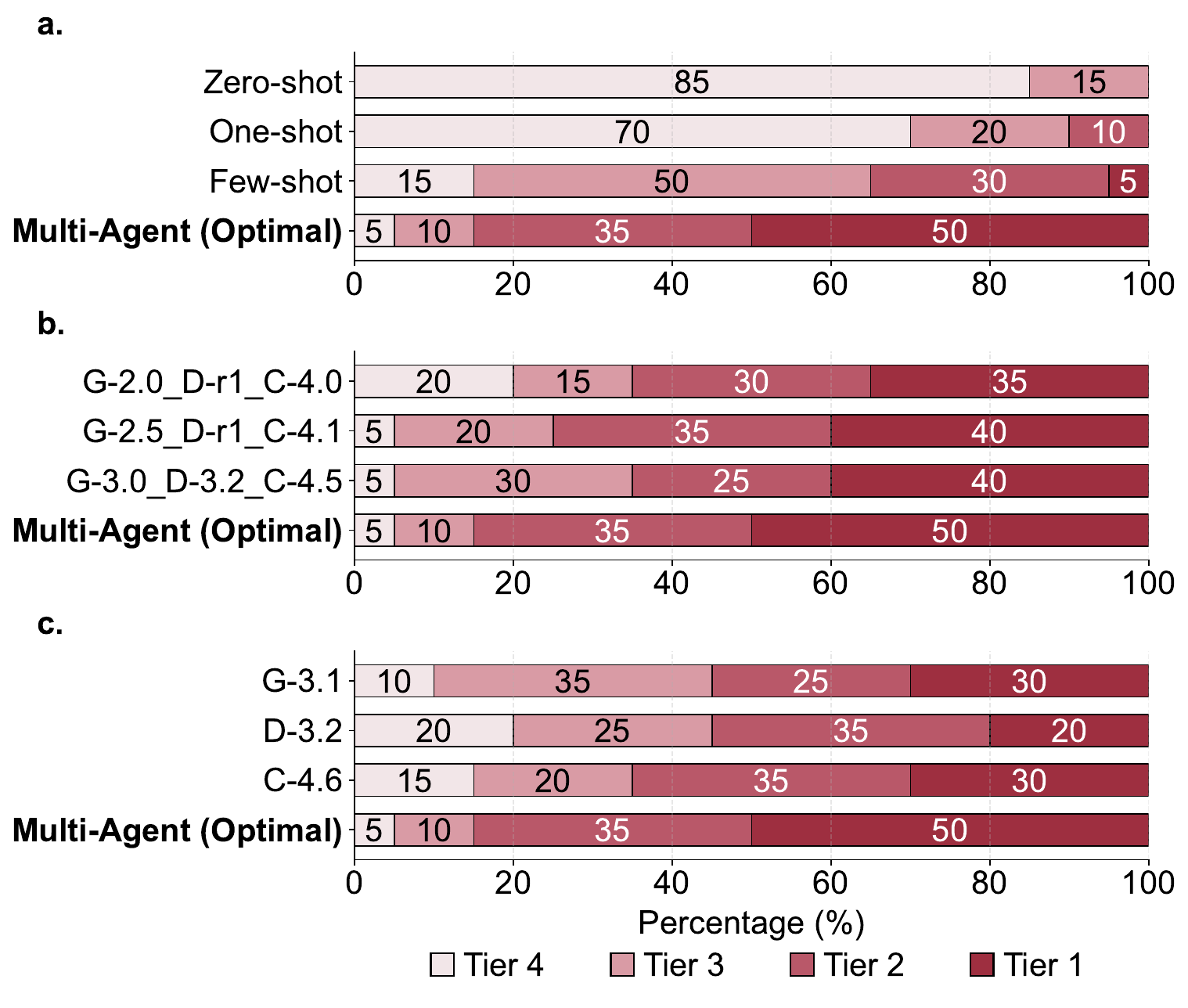}
\caption{\textbf{Quality distribution of autonomously generated compression methodologies.} Performance of 200 synthesized algorithms categorized into four uniform performance tiers (Tier 1 to Tier 4, with Tier 1 representing the highest caliber of innovation). \textbf{a,} Prompting vs. Agentic Workflow. Conventional prompting (using DeepSeek as the baseline, with few-shot examples drawn from seminal literature) predominantly yields physically unviable blueprints. Conversely, the Multi-Agent (Optimal) configuration achieves a 50\% yield of Tier 1 designs. \textbf{b,} Evolutionary scaling of foundation models. Upgrading backbone LLMs (where G, D, and C denote Gemini, DeepSeek, and Claude variants, respectively) within an identical multi-agent architecture demonstrates a clear performance scaling trajectory. \textbf{c,} Multi-agent synergy. Compared to independent single-model agents (G-3.1, D-3.2, C-4.6), the collaborative Multi-Agent (Optimal) framework substantially elevates the proportion of Tier 1 methods from 20–30\% to 50\%, while strictly suppressing structural failures (with Tier 4 designs reduced to a mere 5\%).}
\label{fig5}
\end{figure}

To deconstruct the evolutionary engine’s underlying mechanisms, we conducted extensive ablation studies (Figure~\ref{fig5}). To ensure a rigorous comparison, all generated methods were assessed against a unified global metric derived from the aggregate distribution (see Methods, Automated AI Peer Review Protocol). 

We first evaluated the impact of algorithmic context on generation viability (Figure~\ref{fig5}a). Utilizing DeepSeek as the reasoning baseline, zero-shot generation consistently failed, yielding zero methods in Tier 1 due to severe hardware-incompatible hallucinations. Notably, augmenting the model with highly relevant domain literature (1-shot and few-shot) provided only marginal improvements, indicating that such unguided knowledge injection cannot fundamentally resolve the combinatorial explosion of hardware design. In stark contrast, our Multi-Agent (Optimal) configuration—anchoring generation in an algorithmic CoT—triggered a phase transition in algorithmic reliability, yielding a substantial 50\% share of Tier 1, logically closed blueprints.

Finally, we analyzed the engine's scalability and multi-agent dynamics (Figures~\ref{fig5}b and \ref{fig5}c). 
As the underlying foundation model evolves, the system exhibits a clear overall scaling trajectory, generally expanding the yield of Tier 1 methods (Figure~\ref{fig5}b). Notably, the D-r1 ensemble deviates slightly from this trend, averaging lower than its D-3.2 counterpart. This occurs as D-r1 prioritizes raw reasoning over the conversational compliance and relational alignment typical of highly aligned models~\cite{kalai2026evaluating}. Furthermore, compared to single-model configurations, our heterogeneous multi-agent ensemble serves as a critical quality catalyst (Figure~\ref{fig5}c). While advanced single agents generally hit a capability plateau (hovering at a 20--30\% Tier 1 yield), the generator-reviewer dynamics effectively push promising candidates past the strict acceptance threshold, elevating the Tier 1 yield to 50\% while strictly compressing structural failures (Tier 4) to merely 5\%. 

\section*{Discussion}

This work marks a paradigm shift in the evolution of computational systems: transitioning from manual, heuristic-driven engineering to the directed structural evolution of hardware-aware computing architectures. Traditional automated machine learning approaches have largely remained confined to parametric tuning within predefined search spaces, rendering them incapable of genuine algorithmic discovery. By contrast, our framework demonstrates that machine intelligence can autonomously navigate the intractable combinatorial explosion of hardware--software co-design. By hypothesizing, implementing, and validating novel computational logic, the engine achieves structural parity with expert-level manual designs. This establishes a scalable model for automated architecture design, succeeding precisely where human intuition systematically falters due to the extreme scale and physical complexity of foundation models.

A decisive factor distinguishing this framework from generalist AI agents is the utilization of structured historical memory. The EKG acts as far more than a simple retrieval repository; it serves as a structural prior of human design intelligence, capturing the ``meta-logic'' of successful engineering trajectories. As confirmed by our ablation studies, by explicitly modeling the topological evolution of prior designs and extracting a system-generated algorithmic CoT, the framework bypasses the redundant trial-and-error cycles that plague unconstrained agents. This explicit knowledge transfer provides the necessary physical grounding to transform abstract architectural search into a guided evolutionary process, ensuring that generated blueprints remain strictly compliant with immutable physical limitations.

To close the loop between this conceptual algorithmic discovery and physical instantiation, the framework couples its self-directed logic invention with deterministic execution feedback. Rather than transferring terabytes of raw model weights, the system distills a model's architectural topology and numerical sensitivities into bandwidth-efficient statistical abstractions. This mechanism bridges the critical gap between global topological mutation and local hardware deployment, demonstrating that the full evolutionary cycle of a complex algorithmic blueprint can be executed by machine intelligence in a matter of hours, collapsing a process that traditionally requires months of manual engineering.

Looking ahead, the implications of this discovery engine extend far beyond the compression of foundation models. This work provides a methodological prototype for the autonomous evolution of complex systems characterized by discrete search spaces and rigid physical constraints. In adjacent engineering disciplines—such as Electronic Design Automation (EDA)~\cite{mirhoseini2021graph} and compiler optimization~\cite{mankowitz2023faster}—the fundamental challenge remains the synthesis of optimal computational graphs that map abstract logic perfectly to specific physical substrates. Our framework suggests that the intractable constrained optimization in these fields can be navigated by leveraging an EKG-style repository to formalize the logical rigor of past engineering successes, subsequently extending them via multi-agent logic decomposition.

Despite these advances, the trajectory toward fully autonomous system design faces two critical challenges that delineate future research. First, while the literature ingestion pipeline is highly automated, strict guarantees regarding the structural quality and academic rigor of the EKG currently require minimal human verification. Automating this validation process with high-reliability protocols remains an open challenge for entirely self-contained scientific discovery. Second, the inherent stochasticity of LLM-based reasoning complicates strict algorithmic reproducibility and deterministic performance guarantees at the execution level. Refining the deterministic boundaries of agentic reasoning and developing robust, code-level self-checking mechanisms will be critical to establishing this framework as the foundational infrastructure for future engineering.

Ultimately, this discovery engine addresses a fundamental bottleneck in automated scientific research: traversing high-dimensional hypothesis spaces strictly governed by physical reality. By internalizing the logical rigor of human expertise and executing mathematically complex topological mutations within the unyielding boundaries of the physical world, our framework demonstrates that machine intelligence can evolve from a passive consumer of computational power into an active architect of its own foundations. This establishes a scalable paradigm for machine-driven discovery, paving the way for the automated evolution of self-improving AI architectures.

\clearpage

\clearpage

\newpage
\setcounter{figure}{0}
\renewcommand{\figurename}{Extended Data Figure} 

\section*{Methods}

\subsection*{Construction of the Evolutionary Knowledge Graph}

To initialize the foundational historical repository, we employed a semi-automated retrieval pipeline to curate a domain-specific corpus focused on the algorithmic compression and efficient deployment of large language models (2022--present). Custom scripts programmatically queried and filtered candidate papers from major academic databases based on architectural keywords and venue constraints (restricted to leading computational venues including NeurIPS, ICLR, ICML, CVPR, HPCA, ISCA, and MICRO). The EKG is materialized as a property graph hosted in a Neo4j database to enable highly efficient relational traversals. In this database schema, nodes represent distinct compression methodologies, while edges encode directed relationships (e.g., explicit \textit{Inspiration} and quantitative \textit{Improvement} edges) across methodological trajectories.

\begin{figure}[htb]
\centering
\includegraphics[width=0.75\textwidth]{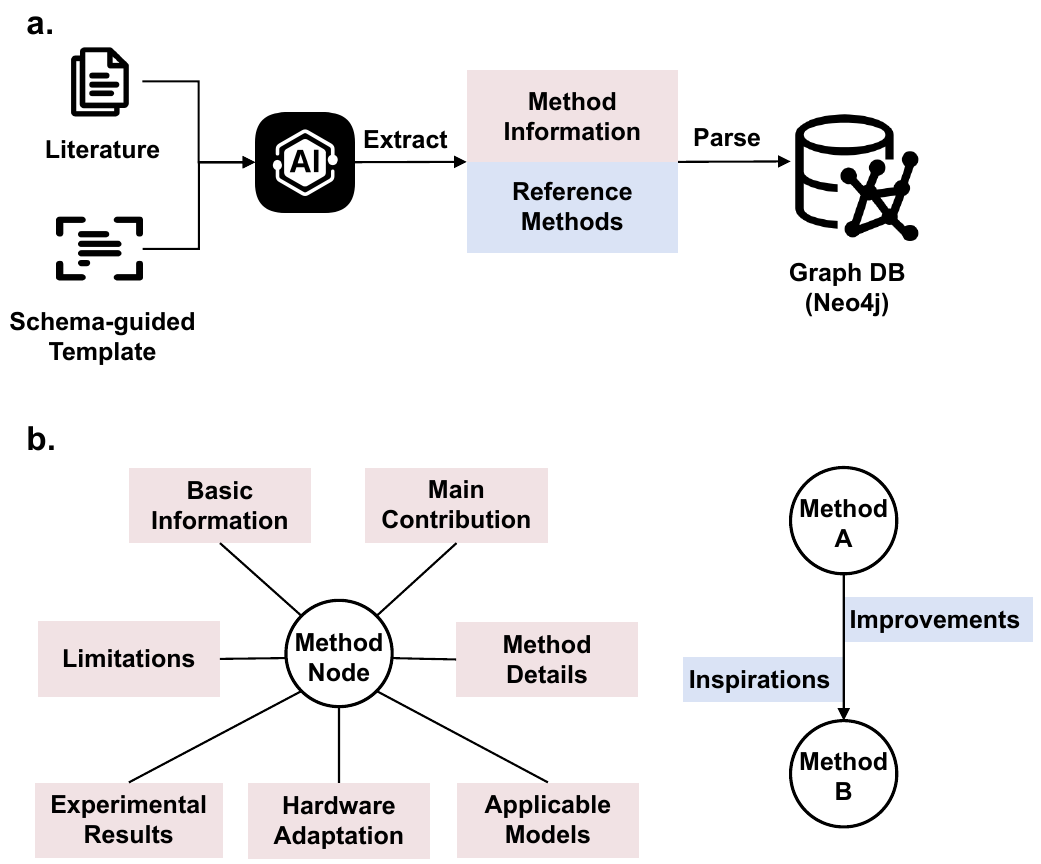}
\caption{\textbf{Construction pipeline and ontology of the EKG.} 
\textbf{a,} Automated graph extraction workflow. Unstructured full-text publications are processed by a schema-guided AI parser utilizing a predefined extraction template. This single-pass neural extraction isolates critical knowledge dimensions, directly parsing the unstructured text into a structured Neo4j graph database. 
\textbf{b,} Node ontology and edge semantics. The graph schema defines formalized metadata representations for each algorithm. A ``Method Node'' explicitly encapsulates seven distinct attribute dimensions: Basic Information, Main Contribution, Method Details, Applicable Models, Hardware Adaptation, Experimental Results, and Limitations. Directed edges structurally connect these nodes, encoding evolutionary relationships such as specific algorithmic \textit{Improvement} (parent-to-child derivations) and conceptual \textit{Inspiration} (analogical similarities).}
\label{fig:ED1}
\end{figure}

\begin{figure}[H]
\centering
\includegraphics[width=0.6\textwidth]{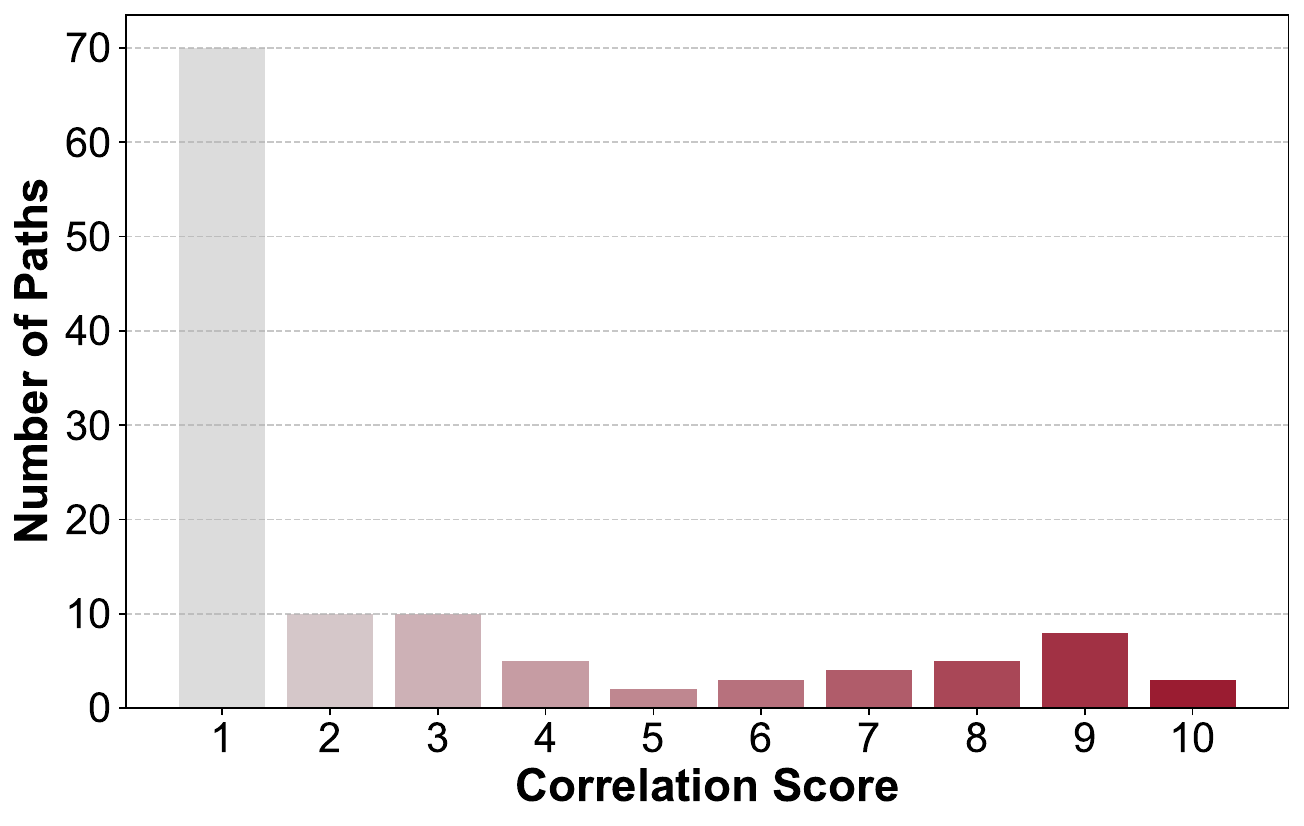}
\caption{\textbf{Negative control evaluation demonstrating hallucination resistance.} Evaluated on a conceptually orthogonal task (``Automated detection and mitigation of memory-resident vulnerabilities''), the engine assigns uniformly low correlation scores (clustering at the lowest bound of 1) to EKG trajectories. This distribution contrasts with the high-score clustering in targeted compression tasks (Fig.~\ref{fig2}b, c). Such a decisive distribution confirms the matching mechanism is strictly task-driven, effectively filtering out incompatible domains and suppressing unconstrained retrieval hallucinations.}
\label{fig:ED5}
\end{figure}

\begin{figure}[H]
\centering
\includegraphics[width=\textwidth]{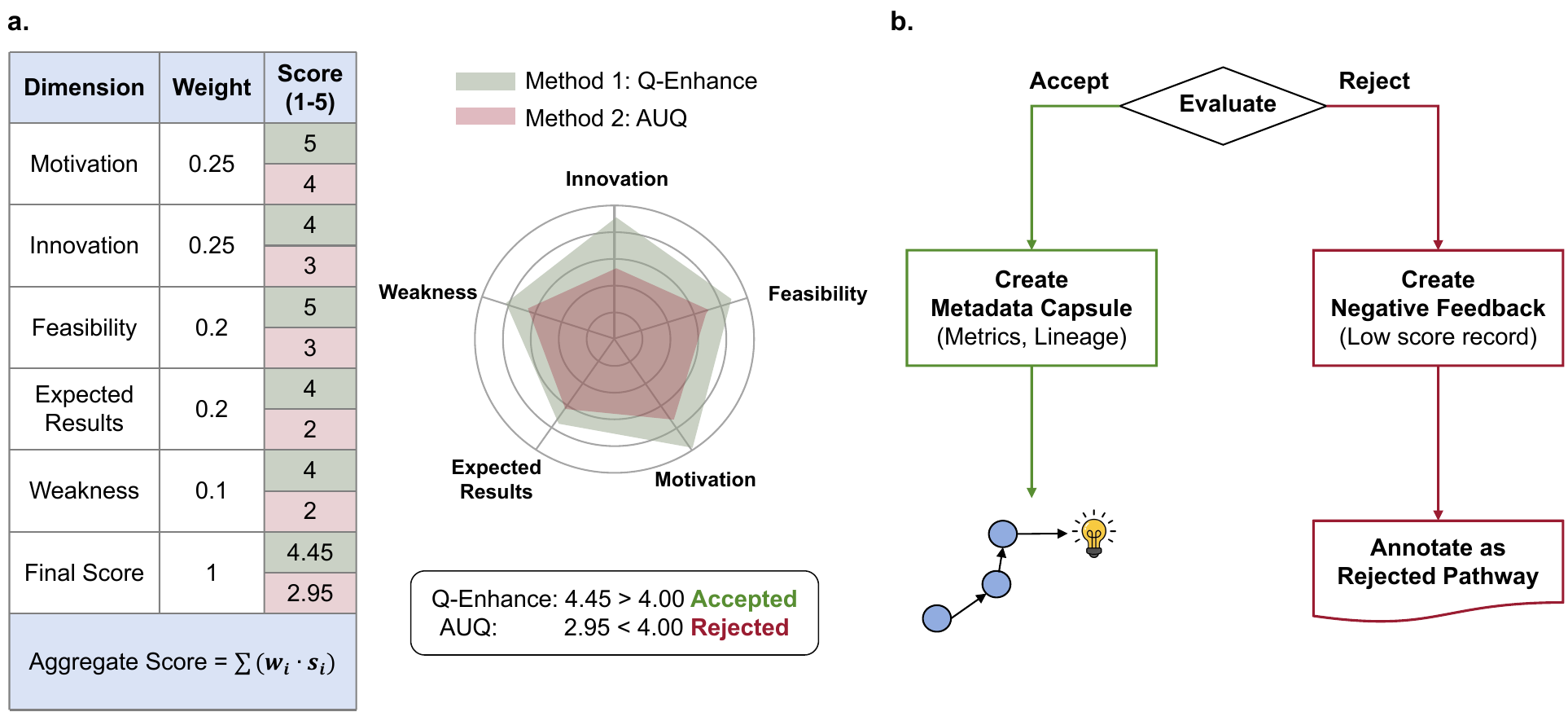}
\caption{\textbf{Quantitative AI peer review rubric and closed-loop evolutionary dynamics.} 
\textbf{a,} Standardization of the automated review process. Candidate blueprints are evaluated on a rigorous 5-point scoring scale across five precisely weighted dimensions. As illustrated in the scoring table and radar chart, the aggregate score computation ($\text{Score}_{\text{agg}} = \sum(w_i \cdot s_i)$) prioritizes Motivation (0.25) and Innovation (0.25) as primary drivers, followed by Feasibility (0.20), Expected Results (0.20), and Weakness (0.10). An explicit acceptance threshold ($\text{Score}_{\text{agg}} \ge 4.00$) is enforced. The visualization contrasts a successful methodology (Method 1: Q-Enhance, score 4.45) against a rejected one (Method 2: AUQ, score 2.95), ensuring that only logically sound designs progress. 
\textbf{b,} Closed-loop feedback mechanism. Based on the evaluation outcome, the system executes a bifurcated knowledge update strategy. Candidates satisfying the threshold (green path) trigger the creation of a Metadata Capsule---capturing critical metrics and structural lineage---which is seamlessly integrated into the EKG as a new conceptual node. Conversely, rejected evaluations (red path) generate Negative Feedback documenting the low-score record. This trajectory is explicitly annotated as a Rejected Pathway, dynamically penalizing its relevance in future traversals to prevent redundant exploration of ineffective methodologies.}
\label{fig:ED2}
\end{figure}

The transformation from unstructured text to the formalized EKG was governed by a neural information extraction pipeline (Extended Data Fig.~\ref{fig:ED1}a). We employed a schema-guided LLM parser alongside a standardized summarization template to decompose documents into formalized metadata representations. Crucially, to ensure strict factual integrity and systematically filter extraction errors and LLM hallucinations, the initial topological mapping and relational edges of this foundational corpus underwent manual auditing. As illustrated in Extended Data Fig.~\ref{fig:ED1}b, each Method Node explicitly encapsulates seven formalized metadata dimensions: (1) Basic Information, tracking origin and provenance; (2) Main Contribution, isolating the core algorithmic innovation; (3) Method Details, documenting specific computational mechanisms and execution logic; (4) Applicable Models, defining algorithmic and architectural compatibility; (5) Hardware Adaptation, specifying target deployment platforms and custom kernel constraints; (6) Experimental Results, storing normalized quantitative performance metrics; and (7) Limitations, establishing methodological boundaries.

Furthermore, the EKG operates as a dynamic repository. During the automated discovery cycle, validated system-evolved methodologies are continuously indexed as new nodes, while empirical failures are appended as trajectory-level metadata. This dynamic graph architecture allows the system to extract complete optimization trajectories, providing the historical context necessary to seed the subsequent multi-agent discovery process. 

\subsection*{Heterogeneous Multi-Agent Evolution}

The multi-agent evolutionary process maps specific tasks to foundation models based on distinct architectural strengths: the Analyzer and Reviewer roles utilize the Gemini Pro series (evaluated across versions 2.5, 3.0, and 3.1) to leverage its native long-context window for deep trajectory dissection and rigorous evaluation; the Ideator is driven by DeepSeek (including R1 and v3.2~\cite{r4}), utilizing its reinforcement-learning-tuned capacity for divergent hypothesis generation; and the Architect role is executed by the Claude Opus family (including versions 4.5 and 4.6)~\cite{r19}, selected for its stringent logical formalization and advanced context-aware code-reasoning capabilities. Finally, the physical code generation is delegated to a dedicated GPT-5.5-based execution agent, which autonomously integrates these verified blueprints with structurally analogous open-source frameworks. This heterogeneous ensemble, encompassing both earlier exploratory checkpoints and the latest stable API releases, serves as the default configuration and was systematically validated via empirical ablation (Figure~\ref{fig5}c).

Prior to generation, an offline computational pre-screening protocol evaluates all EKG paths against target constraints, assigning a relevance score $S \in [1, 10]$ based on methodological advancements and hardware compatibility. To establish a stringent baseline of viability, only trajectories with $S \ge 9$ are retained in a high-potential candidate pool. During the evolutionary cycle, antecedent paths are stochastically sampled from this pool to seed the generation process. The multi-agent ensemble operates with high token efficiency. It typically completes the conceptual evolution of a blueprint within minutes, enabling the end-to-end physical deployment cycle to conclude in a matter of hours.

\subsection*{Automated AI Peer Review Protocol}

Unlike conventional AI reviewers designed to evaluate complete manuscripts, our protocol functions as a dedicated algorithmic verifier. It directly assesses the bare logical implementations and architectural topologies of candidate blueprints. The protocol evaluates these hypotheses on a standardized 5-point scoring scale (Extended Data Fig.~\ref{fig:ED2}a). Aggregate scores ($\text{Score}_{\text{agg}}$) are computed using a rigid weighted rubric: Motivation (0.25), Innovation (0.25), Feasibility (0.20), Expected Results (0.20), and Weakness (0.10). Candidates must satisfy a strict acceptance threshold of $\text{Score}_{\text{agg}} \ge 4.0$. Submissions failing this threshold, or those flagged for mathematical inconsistencies or hardware-incompatible numerics, are preemptively discarded. 

To mitigate LLM-as-a-judge self-preference bias and objectively benchmark our evaluation scale, we executed a controlled empirical benchmarking across $N=25$ algorithms ($N=5$ per category: The AI Scientist~\cite{lu2026aiscientist}, standard venues, preprints, top-tier venues, and our system-evolved blueprints). To ensure zero-shot fairness and prevent data contamination, all human baselines were selected from recent literature strictly withheld from the EKG. Moreover, to eliminate rhetorical bias, the Architect Agent first stripped human papers into bare logical structures, ensuring uniform inputs for the Reviewer. The resulting tier distributions (Extended Data Fig.~\ref{fig:ED4}) confirm the rubric aligns objectively with expert-level human standards, systematically penalizing the topological redundancies and physical infeasibilities inherent to unconstrained generalist agents.

For the ablation studies (Figure~\ref{fig5}), we constructed a master pool of $N=200$ unique generated candidates to mitigate intra-batch variance and score compression. Fixed global tier thresholds were established based on this aggregate distribution, uniformly partitioning the pool into four equi-proportional tiers (Tier 1 to Tier 4). This rigorous stratification ensures that Tier 1 represents a consistent absolute performance benchmark across all experimental axes.

\subsection*{Physical Implementation and Empirical Validation}

To manage computational overhead, only the approved blueprints from the candidate pool are selected for the physical deployment pipeline. Prior to explicit code generation, these blueprints undergo a stringent Pre-Implementation Logic Auditing phase. Driven by the Analyzer agent, this step subjects the conceptual implementations to rigorous semantic and physical constraint checks. Preemptively resolving these latent theoretical flaws allows the system to bypass the massive computational overhead of debugging flawed hardware-level code.

Following this auditing phase, the dedicated execution agent translates the refined designs into executable logic via autonomous algorithmic integration of existing open-source kernels. The derived architectures are then rigorously evaluated on task-specific benchmarks under defined physical constraints. These outcomes dynamically update the EKG: successful architectures are integrated as new nodes, while runtime compilation failures are explicitly annotated as Rejected Pathways (Extended Data Fig.~\ref{fig:ED2}b).

To facilitate the transition from abstract blueprint to tangible execution, this integration phase utilizes a hardware-aware stabilization protocol. By coupling code generation with deterministic execution feedback, the system autonomously calibrates computational logic and resolves architectural bottlenecks in an automated manner. This pipeline enables the execution of complex strategies, including the joint quantization topologies in Q-Enhance and the adaptive routing in MoE-Salient-AQ.

\subsection*{Constraint-Aware Hardware Instantiation}

\begin{figure}[H]
\centering
\includegraphics[width=\textwidth]{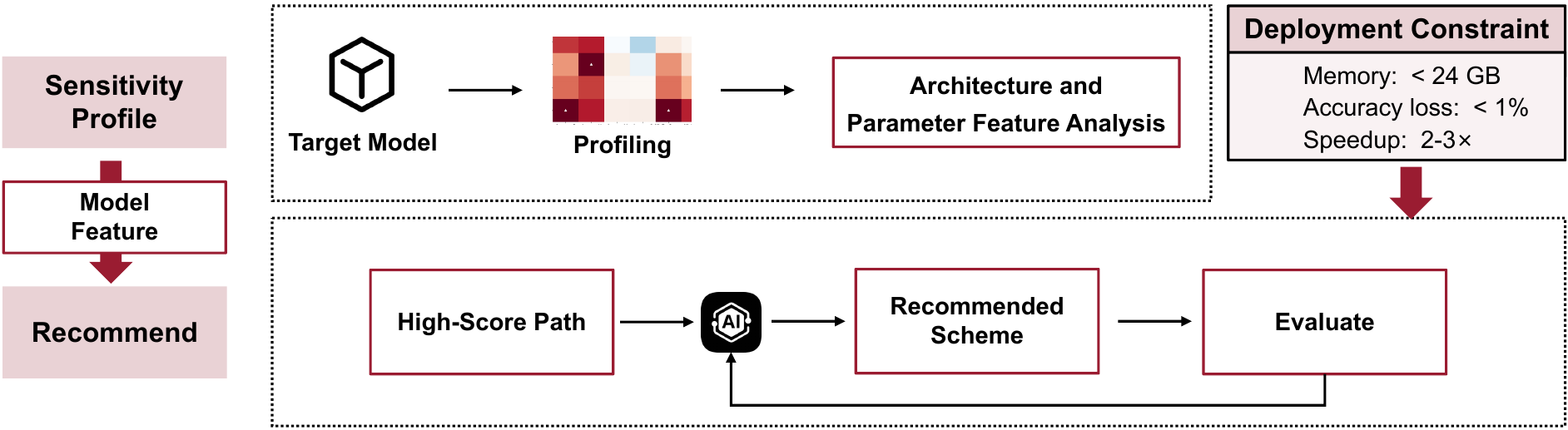}
\caption{\textbf{Hardware-aware recommendation pipeline based on model sensitivity profiling.} The workflow maps the logical transition from sensitivity profiling to iterative hardware-aware recommendation. In the upper panel, a Target Model undergoes initial Profiling (visualized via a heatmap) to drive Architecture and Parameter Feature Analysis, successfully extracting the target's intrinsic Model Features. In the lower recommendation stage, an AI agent synthesizes a Recommended Scheme by integrating three distinct inputs: the extracted model features, explicit Deployment Constraints (e.g., Memory $<$ 24 GB, Accuracy loss $<$ 1\%, Speedup of 2--3$\times$), and a High-Score Path retrieved from the EKG. The proposed scheme then enters the Evaluate module, establishing a continuous feedback loop that routes outcomes back to the AI agent. This dynamic refinement continues iteratively until the generated scheme successfully satisfies all predefined physical deployment boundaries.}
\label{fig:ED3}
\end{figure}

\begin{figure}[H]
\centering
\includegraphics[width=0.9\textwidth]{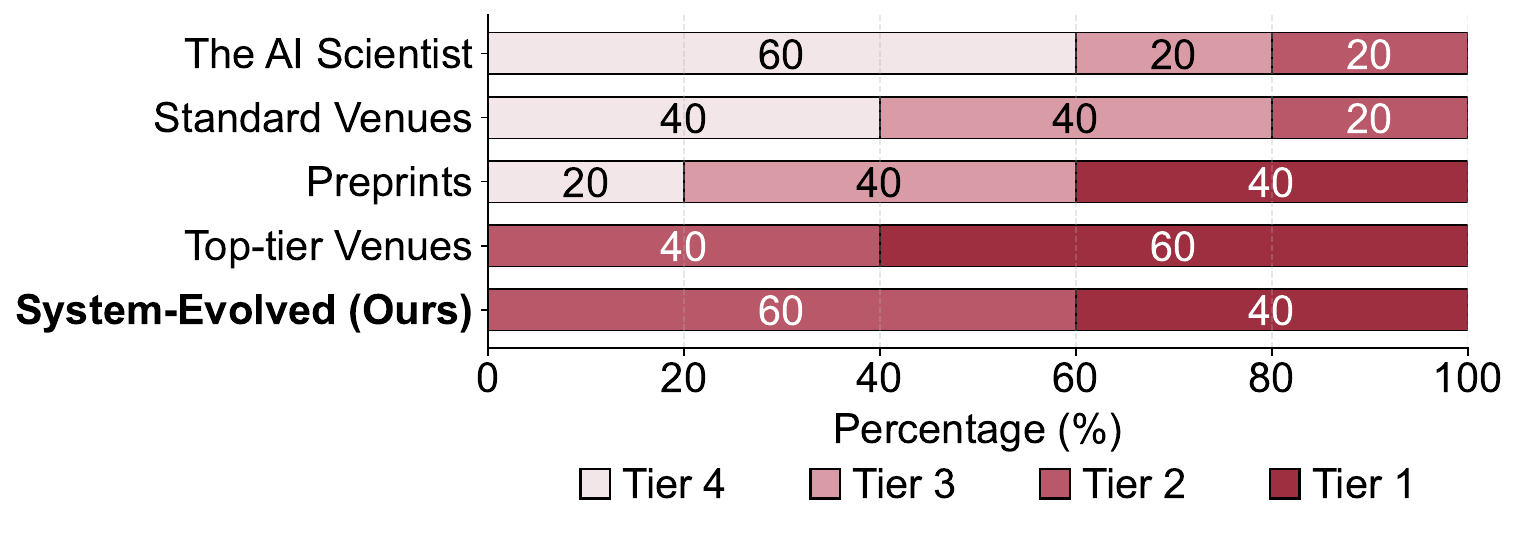}
\caption{\textbf{Quality distribution of compression methodologies across algorithmic provenances.} Percentage breakdown of a representative set ($N=25$ total) of evaluated methodologies, categorized into four uniform performance tiers (Tier 1 to Tier 4, where Tier 1 represents the highest caliber). Generalist AI baselines (The AI Scientist) exhibit unreliability in this domain, with 60\% of their outputs collapsing into Tier 4 and none reaching Tier 1. Methodologies from Standard Venues remain predominantly constrained to the lower-to-middle quality tiers (Tiers 3 and 4). While Preprints demonstrate a higher quality ceiling—with 40\% successfully reaching Tier 1—they exhibit high variance and continue to yield Tier 4 failures. In stark contrast, the System-Evolved (Ours) category---representing the candidate pool that has successfully crossed the autonomous AI peer-review threshold---eliminates sub-optimal outputs. Its structural stability objectively matches the logical rigor of Top-tier Venues, with 100\% of the final synthesized methodologies ranking firmly within the upper two elite tiers (Tiers 1 and 2).}
\label{fig:ED4}
\end{figure}

A dedicated profiler generates the bandwidth-efficient Sensitivity Profile (Extended Data Fig.~\ref{fig:ED3}) locally using 128 calibration sequences sampled from WikiText-2. It outputs a structured JSON payload containing an Architecture Block (mapping layer topologies) and a Numerics Block. The latter characterizes the data distribution of weights and activations, quantifies errors via gradient analysis, and further evaluates the quantization sensitivity of each model component. Crucially, for a representative dense architecture such as Llama-3-8B, this comprehensive payload is condensed to approximately 750 KB. To overcome the inherent numerical reasoning limitations of foundation models when processing dense JSON arrays, these statistical metrics are subsequently rendered into visual trend plots. By simultaneously ingesting both the structured textual payload and these visual representations, the foundation model is empowered to intuitively grasp macro-level phenomena, such as cross-layer outlier distributions and progressive degradation curves. Ultimately, these synthesized model features are integrated with the explicit architectural descriptions to construct the final comprehensive profile that drives the downstream recommendation pipeline.

For hardware instantiation, target environments are mathematically formalized as a constraint tuple $\mathcal{C} = \langle V_{\text{max}}, \Delta A_{\text{max}}, S_{\text{min}} \rangle$, representing the absolute VRAM limit, maximum accuracy degradation, and minimum required speedup, respectively. The Recommender agent queries the EKG using $\mathcal{C}$ and the Numerics Block to initialize a base mixed-precision strategy.

To automatically optimize the recommendation strategy, the Evaluator agent explores the algorithm's discrete architectural space (e.g., bit-width, group size, and symmetry configurations). A deterministic precision-mapping protocol is enforced: structural blocks with high quantization sensitivity receive higher precision, whereas robust layers undergo aggressive quantization. This iterative cycle terminates once the constraints $\mathcal{C}$ are fully satisfied, yielding a standardized configuration file for local compilation.

\subsection*{Experimental Setup and Evaluation Protocol}

To ensure physical reproducibility and support high-throughput evolutionary search, the automated compilation and large-scale evaluation pipelines were deployed on a dedicated hardware node comprising 8$\times$ NVIDIA A100 Tensor Core GPUs (80\,GB, NVLink), elastically supplemented by equivalent cloud computing instances. Although the final large-scale validation utilized this cluster, the multi-agent generation and sensitivity profiling were explicitly restricted to specific LLM hyperparameters and constrained local hardware. As detailed in the empirical results, the framework's generalizability was validated across diverse foundation models (e.g., Llama and Qwen families) using standard reasoning and context benchmarks (e.g., MMLU, LongEval). The efficacy of the agent-derived architectures is benchmarked against canonical post-training quantization and outlier-aware methodologies (e.g., GPTQ~\cite{gptq}, QuaRot~\cite{quarot}, MxMoE~\cite{mxmoe}), with peak memory occupancy precisely measured during active inference and computational speedups calculated relative to uncompressed FP16/BF16 baselines under identical batch-size configurations.

\clearpage

\clearpage

\begin{table}[htb]
\centering
\renewcommand{\tablename}{Extended Data Table}
\caption{\textbf{Performance comparison of the autonomously derived Q-Enhance against baseline quantization methods on the Qwen3 model family.}}
\label{tab:qwen3_qenhance}
\renewcommand{\arraystretch}{1.2}
\setlength{\tabcolsep}{4.5pt} 
\begin{tabular}{ll cccccccc}
\toprule
Models & Methods & Bit & PIQA & ARC-E & ARC-C & H.S & W.G & MMLU & AVG $\uparrow$ \\
\midrule
\multirow{4}{*}{Qwen3-8B} 
 & -           & bf16         & 77.69 & 80.93 & 56.57 & 74.93 & 67.56 & 72.96 & 71.77 \\
 & RTN         & w4kv4        & 74.65 & 75.84 & 52.65 & 72.93 & 64.88 & 68.64 & 68.27 \\
 & AWQ+KVQuant & w4kv4        & 76.39 & 79.38 & 56.40 & 72.87 & 66.06 & 69.43 & 70.09 \\
 & \textbf{Q-Enhance}   & w4.12 kv3.88 & 76.82 & 79.34 & 55.03 & 72.74 & 66.61 & 70.32 & \textbf{70.14} \\
\midrule
\multirow{4}{*}{Qwen3-14B} 
 & -           & bf16         & 79.76 & 82.87 & 60.24 & 78.82 & 72.85 & 77.30 & 75.31 \\
 & RTN         & w4kv4        & 79.33 & 79.92 & 57.00 & 78.06 & 70.48 & 75.37 & 73.36 \\
 & AWQ+KVQuant & w4kv4        & 79.65 & 81.48 & 58.19 & 77.76 & 72.38 & 74.94 & 74.07 \\
 & \textbf{Q-Enhance}   & w3.86 kv4.14 & 79.38 & 81.78 & 59.64 & 77.50 & 72.06 & 76.01 & \textbf{74.40} \\
\bottomrule
\end{tabular}
\end{table}

\begin{table}[htb]
\centering
\renewcommand{\tablename}{Extended Data Table}
\caption{\textbf{Performance comparison of the autonomously evolved MoE-Salient-AQ framework against baseline sparse quantization methods on the Qwen1.5-MoE-A2.7B model.}}
\label{tab:moe_compression}
\renewcommand{\arraystretch}{1.2}
\setlength{\tabcolsep}{4.5pt} 
\begin{tabular}{ll cccccccc c}
\toprule
Methods & Bit & PPL $\downarrow$ & ARC-C & ARC-E & H.S & LO & LS & PIQA & W.G & AVG $\uparrow$ \\
\midrule
- & bf16 & 6.79 & 43.86 & 68.98 & 77.17 & 71.57 & 64.56 & 80.36 & 69.22 & 67.96 \\
\midrule
\multirow{2}{*}{SliM-LLM} 
 & w4a16 & 6.88 & 44.62 & 68.94 & 76.55 & 71.18 & 65.07 & 80.47 & 69.14 & \textbf{68.00} \\
 & w3a16 & 7.27 & 40.36 & 64.86 & 74.94 & 69.47 & 63.79 & 78.94 & 65.75 & 65.44 \\
\midrule
\multirow{4}{*}{MxMoE}
 & w4a16   & 7.10 & 40.87 & 68.06 & 76.21 & 69.69 & 62.33 & 80.69 & 68.90 & 66.68 \\
 & w3.5a16 & 7.09 & 44.03 & 68.01 & 76.29 & 69.18 & 62.95 & 79.49 & 68.11 & 66.87 \\
 & w3a16   & 7.43 & 43.43 & 67.21 & 75.17 & 68.41 & 60.76 & 79.49 & 67.64 & 66.02 \\
 & w2.5a16 & 8.75 & 37.37 & 59.85 & 70.57 & 60.86 & 55.06 & 77.91 & 62.12 & 60.53 \\
\midrule
\multirow{4}{*}{\textbf{MoE-Salient-AQ}}
 & w4a16   & 6.89 & 44.37 & 69.91 & 76.92 & 71.38 & 64.27 & 80.20 & 68.82 & \textbf{67.98} \\
 & w3.5a16 & 6.97 & 45.31 & 70.00 & 76.75 & 70.64 & 63.71 & 80.36 & 68.90 & \textbf{67.95} \\
 & w3a16   & 7.16 & 43.69 & 67.55 & 75.94 & 69.36 & 63.15 & 79.22 & 66.61 & \textbf{66.50} \\
 & w2.5a16 & 7.81 & 40.10 & 62.46 & 73.97 & 67.03 & 60.28 & 79.49 & 63.85 & \textbf{63.88} \\
\bottomrule
\end{tabular}
\end{table}

\newpage
\bibliography{sn-bibliography}

\bmhead{Acknowledgements}
We are grateful to Prof. Guohao Dai for constructive suggestions, and to Zhengzheng Tang and Jidong Chen for their participation in early-stage explorations and insightful discussions. This work was supported by the National Natural Science Foundation of China (Grant No. 62572463).

\bmhead{Data availability}
All baseline foundation models (e.g., LLaMA, Qwen) and standard evaluation benchmarks utilized for empirical validation are publicly accessible via Hugging Face. 

\bmhead{Code availability}
The custom source code are deposited in Zenodo. Upon publication, the complete repository will be made publicly available under an open-source license with a permanent DOI.

\end{document}